\documentclass[10pt,twocolumn,letterpaper]{article}

\usepackage[pagenumbers]{iccv}           %

\usepackage{catchfile} %
\usepackage{adjustbox}
\usepackage{multirow}
\usepackage{colortbl}

\usepackage[accsupp]{axessibility} %

\usepackage{lipsum}  
\usepackage{array}
\usepackage{makecell}
\usepackage{soul}
\newcolumntype{H}{>{\setbox0=\hbox\bgroup}c<{\egroup}@{}}
\usepackage{color}
\usepackage{stfloats}
\definecolor{light}{rgb}{0.5, 0.5, 0.5}

\newcommand{\greenup}{\textcolor{ForestGreen}{$\uparrow$}}
\newcommand{\reddown}{\textcolor{BrickRed}{$\downarrow$}}

\newcommand{\PAR}[1]{\vskip2pt \noindent {\bf #1~}}

\definecolor{iccvblue}{rgb}{0.21,0.49,0.74}
\definecolor{cvprblue}{rgb}{0.21,0.49,0.74}
\definecolor{applegreen}{rgb}{0.05, 0.60, 0.20}
\usepackage[pagebackref,breaklinks,colorlinks,citecolor=iccvblue]{hyperref}

\def\paperID{6} %
\def\confName{ICCV}
\def\confYear{2025}

\title{VFM-UDA++: Improving Network Architectures and Data Strategies for Unsupervised Domain Adaptive Semantic Segmentation}

\author{Brunó B. Englert \quad Gijs Dubbelman \\
Eindhoven University of Technology \\
{\tt\small \{b.b.englert, g.dubbelman\}@tue.nl}
}

\begin{document}
\maketitle
\begin{abstract}
Unsupervised Domain Adaptation (UDA) enables strong generalization from a labeled source domain to an unlabeled target domain, often with limited data. In parallel, Vision Foundation Models (VFMs) pretrained at scale without labels have also shown impressive downstream performance and generalization. This motivates us to explore how UDA can best leverage VFMs. Prior work (VFM-UDA) demonstrated that replacing a standard ImageNet-pretrained encoder with a VFM improves generalization. However, it also showed that commonly used feature distance losses harm performance when applied to VFMs.  Additionally, VFM-UDA does not incorporate multi-scale inductive biases, which are known to improve semantic segmentation. 
Building on these insights, we propose VFM-UDA++, which (1) investigates the role of multi-scale features, (2) adapts feature distance loss to be compatible with ViT-based VFMs and (3) evaluates how UDA benefits from increased synthetic source and real target data.
By addressing these questions, we can improve performance on the standard GTA5 $\rightarrow$ Cityscapes benchmark by +1.4 mIoU. While prior non-VFM UDA methods did not scale with more data, VFM-UDA++ shows consistent improvement and achieves a further +2.4 mIoU gain when scaling the data, demonstrating that VFM-based UDA continues to benefit from increased data availability. The implementation is available at \href{https://github.com/tue-mps/vfm-uda-plusplus}{https://github.com/tue-mps/vfm-uda-plusplus}.

\end{abstract}

\section{Introduction}
\label{sec:intro}

\begin{figure}
    \centering
    \includegraphics[width=0.94\columnwidth]{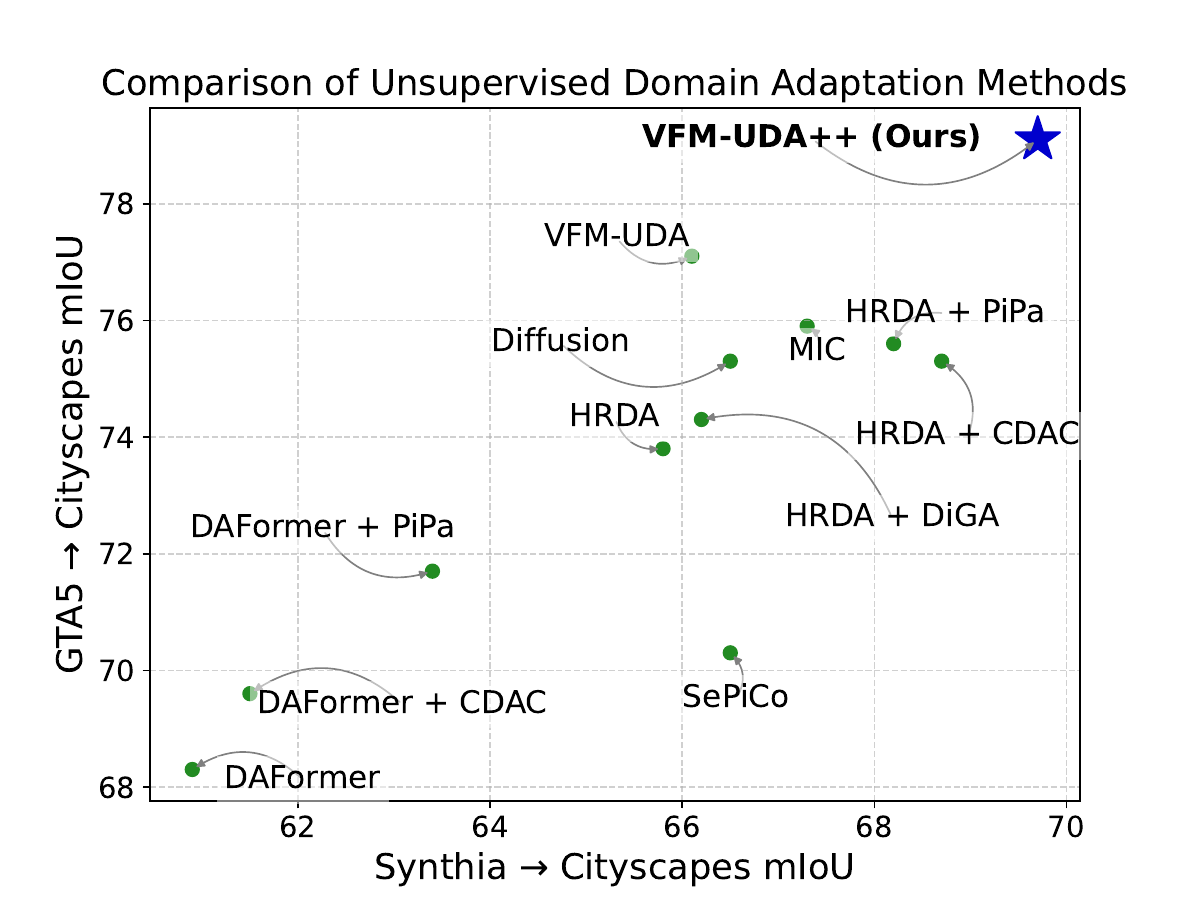}
    \caption{\label{fig:eye-catcher} \textbf{VFM-UDA++ compared to previous state-of-the-art methods}. Result of VFM-UDA++ is shown in blue for the popular benchmarks GTA5 $\rightarrow$ Cityscapes and SYNTHIA $\rightarrow$ Cityscapes.}
    \vspace{-10pt}
\end{figure}

Developing models that adapt to diverse data distributions, including previously unseen distributions, is a key challenge in computer vision, particularly for safety-critical applications like autonomous driving. This challenge is especially pronounced when models encounter significant data distribution shifts between the training and testing conditions.  In dense prediction tasks like semantic segmentation, obtaining accurate pixel-level labeled data is labor-intensive, inherently limiting data scale and diversity. This limitation can hinder adaptation to real-world variability.

Unsupervised Domain Adaptation (UDA)~\cite{Hoffman_featalign_16} aims to bridge the shift in data distribution, often called the \textit{domain gap}, by adapting a model trained on a labeled \textit{source domain} to perform well in an unlabeled \textit{target domain}. This capability is particularly crucial for deploying robust vision systems in autonomous vehicles, which are likely to encounter distribution shifts during real-world deployment. Without adaptation, the domain gap, between source and target caused by variations in location, weather, lighting, textures, sensor properties, class distributions, etc., can lead to overfitting on the source domain and poor generalization to the target domain. UDA techniques have benefited from increasingly powerful vision models, progressing from CNNs pretrained on ImageNet1K~\cite{hoyer_daformer_2022}, multi-scale Vision Transformers (ViTs)~\cite{hoyer_daformer_2022}, to Vision Foundation Models (VFMs)~\cite{englert2024exploring}. These VFMs~\cite{dinov2_2023, eva_2023, eva02_2023}, are pretrained on extensive datasets with weak- and/or self-supervision and exhibit impressive downstream generalization~\cite{dinov2_2023}. Initial work combining VFMs with UDA, namely VFM-UDA~\cite{englert2024exploring}, demonstrated that substituting a traditional ImageNet-pretrained encoder with a VFM significantly improves generalization. However, it also revealed that standard UDA techniques such as feature distance losses do not transfer effectively to the VFM setting.

We build on VFM-UDA~\cite{englert2024exploring} and make two key observations. First, UDA methods using traditional multi-scale architectures (e.g., MIC~\cite{mic_2023}) with ImageNet pretraining achieve strong performance. However, when replacing their multi-scale encoder with a single-scale ImageNet pretrained ViT~\cite{Dosovitskiy_vit_2021}, performance degrades significantly~\cite{englert2024exploring}, which suggests that multi-scale features are necessary for a high performance semantic segmentation. Second, this trend reverses when switching from ImageNet to VFM pretraining. In this case, a single-scale ViT pretrained with self-supervised objectives (e.g., DINOv2~\cite{dinov2_2023}), is able to outperform the multi-scale ImageNet-pretrained architectures, used in prior UDA methods, due to the VFM's large scale pretraining and generalization capabilities. These contrasting observations raise fundamental questions about the influence of the architecture and its pretraining on UDA and whether its generalization can be improved further by combining a multi-scale architecture with strong VFM pretraining. 

Additionally, we revisit the role of data scale in UDA. Recent UDA studies focus on relatively small-scale data settings, but the potential of UDA is to leverage large quantities of unlabeled data. This is especially relevant in autonomous driving, where synthetic driving scenes are increasingly available at scale, and vehicles naturally collect large volumes of unlabeled real-world data during operation. This makes UDA particularly well-suited to benefit from increased data scale. Recent work on source-only fine-tuning with VFMs~\cite{wei2024stronger} shows that performance improves with more synthetic source data, even without using the target domain. This further supports the idea that UDA, which has access to both synthetic source and real target data, could benefit from scaling data.

Thus, this work addresses the following research questions: (i) Can ViT-based VFM UDA benefit from reintroducing multi-scale inductive biases? (ii) Can feature distance losses be adapted to work with VFMs? (iii) Does scaling synthetic source and unlabeled real target data improve VFM-based UDA performance, and how should such data be composed?

To this end, we propose \textbf{VFM-UDA++}, a method that extends VFM-UDA to investigate and address these questions. We add ViT-Adapter~\cite{chen2022vitadapter} to reintroduce multi-scale features into VFM based UDA and evaluate a range of decoder designs. We also develop a feature distance loss that is compatible with VFM-based encoders—specifically, it avoids class-based masking and can be applied without labels on both source and target domains. We also explore the effects of scaling synthetic source and unlabeled target data in a unified setting.

Our findings demonstrate that: (i) multi-scale features improve UDA performance, but only when paired with appropriate decoder architectures, where simpler, well-matched decoders can outperform more complex ones; (ii) feature distance losses remain useful for ViT-based VFMs when properly adapted to the pretrained representation; and (iii) VFM-UDA++ continues to improve with increased data, whereas prior non-VFM methods plateau or degrade, highlighting a significant opportunity for scaling UDA with synthetic and unlabeled real data.

These insights establish VFM-UDA++ as a targeted refinement of VFM-UDA, providing architectural and empirical clarity on how to better adapt Vision Foundation Models for Unsupervised Domain Adaptation.

\section{Related Work}
\label{sec:relwork}
Our contributions build on a significant body of previous UDA work~\cite{wang_domain_2021,zhang_prototypical_2021,hoyer_daformer_2022,Wang2023CDACCA,Chen_pipa_2023,hoyer_hrda_2022,shen2023diga,PengDiffusion,mic_2023,chen2024transferring,englert2024exploring}. Over years of research, \textit{pseudo-labeling}~\cite{laine2017temporal, araslanov_self-supervised_2021, hoyer_hrda_2022, mic_2023, cluda_2022, xie_sepico_2022} emerged as the core (self-supervised) UDA technique, using target domain predictions as pseudo-labels to guide adaptation, often stabilized through Exponential Moving Average (EMA) models. Effective \textit{learning rate optimization} strategies, such as using different rates for encoder and decoder layers and employing learning rate warm-up phases~\cite{hoyer_daformer_2022}, have proven critical for preserving robust pretrained features while fine-tuning for downstream tasks. \textit{Data augmentation} and source-target sample mixing~\cite{tranheden_dacs_2021} expose models to diverse spatial and semantic contexts, promoting robustness to pixel-space differences. \textit{Architectural enhancements}, including multi-scale encoders and attention modules~\cite{hoyer_daformer_2022, hoyer_hrda_2022}, are designed to capture better fine details and contextual information necessary for accurate dense predictions. To tackle \textit{class imbalance}, methods like Rare Class Sampling~\cite{hoyer_daformer_2022} mitigate issues in poorly balanced datasets, reducing the risk of class prediction collapse. \textit{Feature distance loss}~\cite{hoyer_daformer_2022} aligns source and/or target feature distributions by minimizing discrepancies between their feature representations. Finally, \textit{masked image consistency}~\cite{mic_2023} techniques enhance contextual reasoning by enforcing consistent predictions across masked and unmasked views.

A modern and complementary non-UDA technique to improve model generalization is using Vision Foundation Model (VFM) pre-training. VFMs have significantly advanced computer vision by leveraging large-scale, diverse data for versatile downstream applications. For instance, CLIP~\cite{radford_clip_2021} uses contrastive learning~\cite{Chen_contlearn_2020} on image-text pairs for robust visual representations, while MAE~\cite{mae_2022} reconstructs pixel-level data through masked image modeling~\cite{Xie_simmim_22}. DINOv2~\cite{dinov2_2023}, a recent self-supervised method, achieves strong cross-domain generalization without labeled data. However, most VFMs rely on single-scale ViT~\cite{Dosovitskiy_vit_2021} architectures, which can be less suitable for dense tasks like semantic segmentation. Recent efforts, such as the ViT-Adapter~\cite{chen2022vitadapter}, address this by enabling multi-resolution feature output. Initial research combining VFMs with UDA was conducted in~\cite{englert2024exploring} but did not include contemporary multi-resolution methods such as the ViT-Adapter.

\section{Methodology}
\label{sec:method}
VFM-UDA++ builds on top of VFM-UDA~\cite{englert2024exploring}, aiming to address three open questions: (i) can multi-scale inductive biases improve VFM-based UDA? (ii) can feature distance losses be made effective for ViT-based VFMs? and (iii) how does UDA performance evolve with large-scale synthetic and real data?

\subsection{VFM-UDA++}
\begin{figure}
    \centering
    \includegraphics[width=0.75\linewidth]{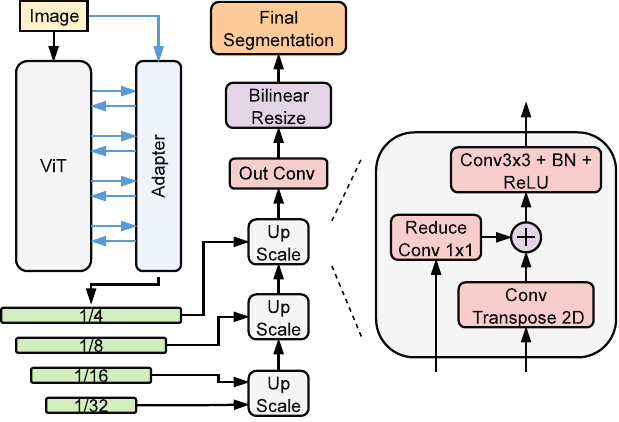}
    \caption{\label{fig:main_arch} \textbf{VFM-UDA++ multi-scale architecture}. VFM-UDA++ adds multi-scale inductive bias to VFM-UDA by equipping it with a ViT-Adapter~\cite{chen2022vitadapter} and a custom image pyramid decoder \textit{BasicPyramid}.}
    \vspace{-10pt}
\end{figure}

As shown in \cref{fig:main_arch}, VFM-UDA++ combines a ViT-based encoder pretrained with DINOv2, a ViT-Adapter to produce multi-scale features, and our custom BasicPyramid decoder. Its training strategy extends VFM-UDA by introducing a feature distance loss tailored for ViT-based VFMs.

\PAR{Core UDA techniques:} 
VFM-UDA++ builds on pseudo-labeling, a widely used approach in UDA~\cite{laine2017temporal, araslanov_self-supervised_2021, hoyer_hrda_2022, mic_2023, cluda_2022, xie_sepico_2022}. To improve its robustness, we incorporate an Exponential Moving Average (EMA) teacher model~\cite{tarvainen_mean_2017} to generate more stable pseudo-labels across training iterations. Additionally, we apply pseudo-label weighting~\cite{french_semi-supervised_2020, Olsson2021WACV} to prioritize high-confidence predictions and suppress noisy ones, particularly during the early training stages, when pseudo-label accuracy is low.

\PAR{Multi-scale VFM encoder:} Vision Transformers (ViTs) typically output single-scale features, which can be limiting for semantic segmentation tasks that benefit from multi-scale, high-resolution features. Unlike VFM-UDA~\cite{englert2024exploring}, we address this by extending ViT-based VFMs~\cite{dinov2_2023} with a ViT-Adapter~\cite{chen2022vitadapter} that outputs multi-resolution features, enabling accurate segmentation for object edges and small objects. Our experiments primarily utilize DINOv2 as the ViT-based VFM due to its robust performance, as shown in ~\cref{tab:vfmpretraining}.

\PAR{Multi-scale decoder:}
To leverage the multi-resolution features produced by the ViT-Adapter, we pair it with our custom \textit{BasicPyramid} decoder. This decoder is lightweight by design, using only standard convolutions~\cite{lecun2015deep}, ReLU~\cite{agarap2018relu}, and BatchNorm~\cite{szeged2015batchnorm}. Unlike DAFormer~\cite{hoyer_daformer_2022}, which upsamples all encoder features to a single high-resolution scale, incurring significant memory and compute overhead, our decoder operates on a true feature pyramid. This structure is widely used and substantially more resource efficient, as demonstrated in \cref{sec:results_multiscalearch}.

Compared to prior work such as HRDA~\cite{hoyer_hrda_2022} or MIC~\cite{mic_2023}, which require multiple forward passes during inference for a single image, our design processes the full-resolution input in one forward pass. We explored other decoder types, including transformer-based alternatives, but found them computationally more expensive and not consistently more effective. Despite its simplicity, BasicPyramid achieved the highest segmentation accuracy among the decoders we evaluated, while offering much better computational efficiency. This makes it the preferred decoder for VFM-UDA++.

\PAR{Feature Distance Loss:}
Due to ImageNet~\cite{Russakovsky_imagenet_2015} dataset's label bias toward thing classes, ImageNet pretrained models do rarely observe stuff-classes. To resolve this bias, DAFormer~\cite{hoyer_daformer_2022} introduced a feature distance loss specifically for ImageNet pretrained models to align student features with those of an ImageNet-pretrained backbone. However, this FD loss can only be applied on labeled data and thus cannot be used on the unlabeled target domain. Later work~\cite{englert2024exploring} also found that such masking based FD losses can degrade performance when used with ViT-based Vision Foundation Models (VFMs). In VFM-UDA++, we propose a VFM-compatible FD loss that addresses both issues. VFMs like DINOv2 are pretrained on large-scale, diverse data with a self-supervised loss. Thus by using a VFM as the teacher allows us to apply a feature distance loss without masking. Since labels are no longer required, we can apply a feature distance loss on both the source and target domains. 

The proposed feature distance loss computes a weighted combination of cosine similarity and $\ell_1$ distance between teacher and student features:

\begin{equation*}
\mathcal{L}_{\text{FD}} = \lambda_{\text{cos}} \cdot \left(1 - \cos(f^T, f^S)\right) + \lambda_{\ell_1} \cdot \|f^T - f^S\|_1,
\end{equation*}

where $f^T$ are that features from the teacher and $f^S$ are the projected features from the student model.

\begin{figure}
    \centering
    \includegraphics[width=0.9\linewidth]{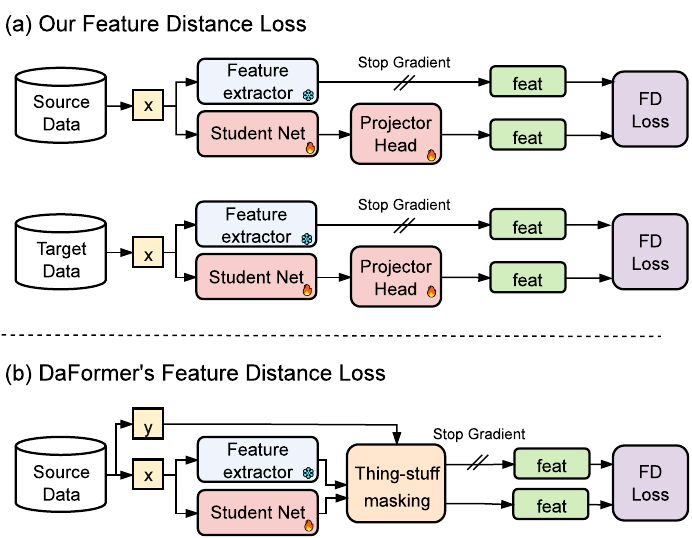}
    \caption{\label{fig:fd_loss} \textbf{Our proposed feature distance loss compared to that of DAFormer}. Our VFM Feature distance loss works on both source and target data. This is different from the DAFormer~\cite{hoyer_daformer_2022} feature distance loss, where labels are a crucial part of the FD loss and thus can only be used for the source data where labels are available.}
    \vspace{-10pt}
\end{figure}

As illustrated in \cref{fig:fd_loss}, this formulation enables fully label-free application of the FD loss across both domains, without masking or requiring labeled data.

\PAR{Implementation details:}
The VFM-UDA++ architecture is a ViT encoder pretrained with DINOv2 coupled with ViT-Adapter~\cite{chen2022vitadapter} and the BasicPyramid decoder. Training uses the AdamW~\cite{loshchilov2018adamw} optimizer, with a linear warm-up learning rate for the first 1,500 iterations, followed by linear decay. The decoder learning rate was set at 1.4$\times10^{-4}$, while the encoder used a learning rate of 1.4$\times10^{-5}$ and the DINOv2 pretrained ViT's layers have 0.9 layerwise decay. In all experiments, no matter the dataset, the training iteration is 40,000 with a batch size of 8. For pseudo-label generation, we use horizontal flip aggregation to minimize pseudo-label noise and the threshold for accepted pseudo-labels is 0.968~\cite{mic_2023,hoyer_daformer_2022}. We use image masking with a 0.7 drop rate, the same as in MIC~\cite{mic_2023}. Additionally, during training cross-domain mixed sampling (DACS)~\cite{tranheden_dacs_2021} is used to enhance domain robustness, and rare class sampling~\cite{hoyer_daformer_2022} to improve representation of underrepresented categories.

\subsection{Experimental Setup}
\label{sec:exp_setup}
To evaluate UDA in a setting more reflective of the challenges in autonomous driving, we adopt an experimental design that emphasizes data diversity, scale, and domain generalization. Specifically, we (i) introduce a large-scale evaluation protocol to study scalability, (ii) distinguish between in-target and out-of-target generalization, and (iii) define a new cross-domain oracle to better estimate the attainable upper bound for UDA methods.

\PAR{Large-scale data setup:}
We construct the synthetic source domain (All-synth) by combining GTA5~\cite{richter_playing_nodate}, Synthia~\cite{ros2016synthia}, and Synscapes~\cite{Wrenninge2018Synscapes}, totaling 59,367 labeled images. For the unlabeled target domain (All-real), we combine Cityscapes~\cite{cordts_cityscapes_2016}, BDD100K~\cite{yu2020bdd100k}, Mapillary Vistas~\cite{Neuhold_mapillary_17}, ACDC~\cite{Sakaridis2021acdc}, and DarkZurich~\cite{dai2018darkzurich}, for a total of 41,356 real-world images. This All-synth → All-real configuration enables evaluation under high variability in lighting, geography, and scene structure.

\PAR{In-target vs. out-of-target evaluation:}
Following~\cite{englert2024exploring}, we evaluate performance on two axes. \textit{In-target} evaluation measures performance on the same target domain used during UDA training. Specifically, we evaluate on the validation set of the target data. This provides insight into how well the model adapts to our target data domain. This measurement is always done in UDA benchmarks. \textit{Out-of-target} evaluation tests generalization to never seen domains. We use WildDash2~\cite{Zendel_wilddash_18}, which was not used during training, but belongs to the same broader domain of real-world driving scenes. WildDash2 includes more diverse geographies, lighting conditions, and sensor types, making it a challenging and realistic test for out-of-domain generalization. Together, these metrics assess both the model’s alignment with the adapted target domain and model's ability to generalize across domains.

\PAR{Cross-domain oracle:}
To complement standard evaluation, we introduce a \textit{cross-domain oracle}, a new concept designed to establish a more realistic upper bound for UDA generalization. The motivation arises from qualitative observations, where VFM-UDA++ occasionally produces predictions that appear more accurate than the Cityscapes ground truth (\cref{fig:qualitative}). Prior work~\cite{hoyer_daformer_2022} has also documented substantial labeling inconsistencies across datasets. Such inconsistencies can affect both training and evaluation, introducing an artificial ceiling on reported UDA performance.

To better account for this, we define the cross-domain oracle as a model trained on all labeled datasets within \textit{All-real}, explicitly excluding Cityscapes, but evaluated on the Cityscapes validation set. This setup avoids any use of target-domain during training, while still providing diverse training dataset. In theory, if annotation protocols were consistent across datasets, this model should perform comparably to a standard oracle trained directly on Cityscapes. However, we observe a notable performance gap between the two, suggesting that the remaining performance difference may be due not only to domain shift but also to label noise or annotation mismatches.

By comparing UDA performance to this cross-domain oracle, we gain a clearer understanding of whether the gap to oracle performance stems primarily from domain shift or from deeper inconsistencies in the benchmarks. As such, the cross-domain oracle provides a more realistic and informative upper bound than the standard oracle used in previous works.

\section{Experiments}
\label{sec:results}

We evaluate VFM-UDA++ across a range of experimental setups, structured around four main investigations: (1) performance on standard UDA benchmarks, (2) the effect of reintroducing multi-scale inductive biases, (3) the role of adapted feature distance losses for ViT-based VFMs, and (4) how UDA performance scales with additional synthetic source and unlabeled target data. We further examine how different dataset compositions affect adaptation, analyze in-target and out-of-target generalization, and compare against various oracle models

\subsection{Results on standard UDA benchmarks}
We first evaluate VFM-UDA++ on widely used UDA benchmarks, including GTA5 $\rightarrow$ Cityscapes, Synthia $\rightarrow$ Cityscapes, and Cityscapes $\rightarrow$ DarkZurich.  As demonstrated by ~\cref{tab:gta2cs} and \cref{tab:extraresults}, there is consistent improvement over prior methods, where VFM-UDA++ outperforms both VFM-UDA and non-VFM UDA methods across all settings.
Notably, VFM-UDA++ S (27M parameters) achieves 76.6 mIoU on GTA5 $\rightarrow$ Cityscapes, surpassing the best-performing non-VFM model (MIC~\cite{mic_2023}) by +0.7 mIoU, despite having over 3$\times$ fewer parameters. 

\begin{table}[tb] 
\centering 
\begin{adjustbox}{width=0.96\linewidth}
\begin{tabular}{lccccc} 
\toprule
\textbf{Method} & \textbf{$\#$Parameters} & \textbf{Backbone} & \textbf{mIoU CS} \\ 
\midrule
CorDA~\cite{wang_domain_2021} \textbackslash w depth   & 48M  & ResNet-101     & 56.6 \\
ProDA~\cite{zhang_prototypical_2021}                   &  48M  & ResNet-101    & 57.5 \\
SePiCo~\cite{xie_sepico_2022}                          &  86M  & MiTB-5        & 70.3 \\
DAFormer~\cite{hoyer_daformer_2022}                    &  86M  & MiTB-5        & 68.3 \\
DAFormer~\cite{hoyer_daformer_2022}~+~CDAC~\cite{Wang2023CDACCA} &  86M  & MiTB-5 & 69.6 \\
DAFormer~\cite{hoyer_daformer_2022}~+~PiPa~\cite{Chen_pipa_2023} &  86M  & MiTB-5 & 71.7 \\
HRDA~\cite{hoyer_hrda_2022}                            &  86M  & MiTB-5        & 73.8 \\
HRDA~\cite{hoyer_hrda_2022}~+~DiGA~\cite{shen2023diga} &  86M  & MiTB-5        & 74.3 \\
HRDA~\cite{hoyer_hrda_2022}~+~CDAC~\cite{Wang2023CDACCA} &  86M  & MiTB-5        & 75.3 \\
Diffusion~\cite{PengDiffusion}                         &  86M  & MiTB-5        & 75.3 \\
HRDA~\cite{hoyer_hrda_2022}~+~PiPa~\cite{Chen_pipa_2023}  &  86M  & MiTB-5        & 75.6 \\
MIC~\cite{mic_2023}                                    &  86M  & MiTB-5        & 75.9 \\
DCF~\cite{chen2024transferring} \textbackslash w depth & 86M  & MiTB-5         & 77.7 \\
\midrule
VFM-UDA~\cite{englert2024exploring}          &  27M  & DINOv2-S         & 71.3 \\
VFM-UDA~\cite{englert2024exploring}          &  88M  & DINOv2-B         & 77.1 \\
VFM-UDA~\cite{englert2024exploring}         &  334M & DINOv2-L         & 78.4 \\
\midrule
\textcolor{gray}{Source-only} &  \textcolor{gray}{334M}  & \textcolor{gray}{DINOv2-L}      & \textcolor{gray}{71.8} \\
VFM-UDA++  &  27M   & DINOv2-S      & \textbf{76.6}~\greenup~\textcolor{applegreen}{+5.3}  \\
VFM-UDA++  &  88M   & DINOv2-B      & \textbf{79.1}~\greenup~\textcolor{applegreen}{+2.0}  \\
VFM-UDA++  &  334M  & DINOv2-L      & \textbf{79.8}~\greenup~\textcolor{applegreen}{+1.4}  \\
\textcolor{gray}{Oracle} &  \textcolor{gray}{334M}  & \textcolor{gray}{DINOv2-L}      & \textcolor{gray}{85.0} \\
\bottomrule
\end{tabular}
\end{adjustbox} 
\caption{\label{tab:gta2cs} \textbf{GTA5 $\rightarrow$ Cityscapes benchmark}. VFM-UDA++ consistently outperforms previous methods. The provided deltas for VFM-UDA++ are to VFM-UDA for the same model size.}

\vspace{-10pt} 
\end{table}

\begin{table}[tb]
\centering 
\begin{adjustbox}{width=1.0\linewidth}
\begin{tabular}{lcccc}
\toprule 
\textbf{Method} & \textbf{Backbone} & \textbf{Source} & \textbf{Target}  &  \textbf{mIoU}  \\
\midrule
ProDA~\cite{zhang_prototypical_2021}  & ResNet-101 & Synthia & Cityscapes & 55.5 \\
DAFormer~\cite{hoyer_daformer_2022} & MiTB-5 & Synthia & Cityscapes  &  60.9 \\
DAFormer~\cite{hoyer_daformer_2022}~+~CDAC~\cite{Wang2023CDACCA} & MiTB-5 & Synthia & Cityscapes  &  61.5 \\
DAFormer~\cite{hoyer_daformer_2022}~+~PiPa~\cite{Chen_pipa_2023} & MiTB-5 & Synthia & Cityscapes  &  63.4 \\
HRDA~\cite{hoyer_hrda_2022} & MiTB-5 & Synthia & Cityscapes  & 65.8 \\
VFM-UDA~\cite{englert2024exploring} & DINOv2-B &  Synthia & Cityscapes & 66.1  \\
HRDA~\cite{hoyer_hrda_2022}~+~DiGA~\cite{shen2023diga} & MiTB-5 &  Synthia & Cityscapes & 66.2 \\
Diffusion~\cite{PengDiffusion} & MiTB-5 & Synthia & Cityscapes  & 66.5 \\
SePiCo~\cite{xie_sepico_2022} & MiTB-5 & Synthia & Cityscapes & 66.5 \\
MIC~\cite{mic_2023} & MiT-B5 & Synthia & Cityscapes & 67.3 \\
HRDA~\cite{hoyer_hrda_2022}~+~PiPa~\cite{Chen_pipa_2023}  & MiTB-5 & Synthia & Cityscapes  & 68.2 \\
HRDA~\cite{hoyer_hrda_2022}~+~CDAC~\cite{Wang2023CDACCA} & MiTB-5 & Synthia & Cityscapes  & 68.7 \\
VFM-UDA++ & DINOv2-B &  Synthia & Cityscapes & \textbf{69.7}~\greenup~\textcolor{applegreen}{+1.0} \\

\midrule
SePiCo~\cite{xie_sepico_2022} & MiTB-5 & Cityscapes & Darkzurich & 45.4 \\
MIC~\cite{mic_2023} & MiT-B5 & Cityscapes & Darkzurich & 60.2  \\
VFM-UDA~\cite{englert2024exploring} & DINOv2-B &  Cityscapes & Darkzurich & 67.4  \\
VFM-UDA++ & DINOv2-B & Cityscapes & Darkzurich & \textbf{68.7}~\greenup~\textcolor{applegreen}{+1.3} \\

\bottomrule
\end{tabular}
\end{adjustbox}
\caption{\textbf{Other popular UDA benchmarks}. VFM-UDA++ consistently outperforms prior methods across all benchmarks including challenging settings like Cityscapes-to-Darkzurich and Synthia-to-Cityscapes. } 
\label{tab:extraresults}

\vspace{-10pt}
\end{table}

\subsection{Effect of Multi-Scale Encoders and Decoders}
\label{sec:results_multiscalearch}

\PAR{Ablation of multi-scale encoder:} We start with choosing the VFM encoder and assess single-scale and multi-scale options. The benefit of a multi-scale VFM lies in the ease of applying multi-scale inductive bias for UDA, while for single-scale VFMs additional components (e.g.,~ViT-Adapter~\cite{chen2022vitadapter}) are required. As shown in ~\cref{tab:vfmpretraining}, the single-scale VFM DINOv2~\cite{dinov2_2023} achieves the highest performance, with an in-target mIoU of 80.3\% and an out-of-target mIoU of 68.6\%, making it the preferred choice for VFM-UDA++. While the multi-scale VFMs are competitive and better than their non-VFM counterparts, their generalization is not on par with that of the single-scale DINOv2. %

This brings us to the next ablation, where we assess if adding multi-scale inductive biases to a single-scale VFM encoder helps UDA. For this, we test several encoder configurations as shown in ~\cref{tab:multiscale}. Starting with a baseline method VFM-UDA~\cite{englert2024exploring} with a ViT-B/14~\cite{Dosovitskiy_vit_2021} DINOv2~\cite{dinov2_2023} pretrained encoder and a VFM-UDA decoder~\cite{englert2024exploring}, we achieve an mIoU of 79.1\%, which serves as the reference. Adding the ViT Adapter improves mIoU to 79.9\% (+0.8 mIoU). However, adding two extra ViT Blocks without multi-scale capabilities—solely to match the parameter count of the ViT Adapter—slightly decreases performance to 78.8\% (-0.3 mIoU). This shows that the multi-scale capability and the inductive bias of the ViT-Adapter help in UDA.

\begin{table}[tb]
\centering
\begin{adjustbox}{width=0.95\linewidth}
    \begin{tabular}{lcccc}
    \toprule
        \textbf{Pre-training method} & \textbf{VFM}  & \textbf{Multi-scale}  &\thead{\textbf{In-target} \\  \textbf{mIoU CS}} & \thead{\textbf{Out-of-target} \\ \textbf{mIoU WD2}}  \\ 
        \midrule
        Deit3 IN1k~\cite{deit3_22}           & - & - & 72.5 &  58.6 \\
        MAE~\cite{mae_2022}                  & - & - & 73.5 &  54.2 \\ 
        Deit3 IN22k~\cite{deit3_22}          & - & - & 73.7 &  61.7 \\
        SwinV2~\cite{liu2021swinv2}          & - & \checkmark  & 75.1 &  61.5 \\
        EVA-02~\cite{eva02_2023}             & \checkmark & -  & 75.5 & 66.9 \\ 
        EVA-02-CLIP~\cite{Sun_eva02clip_23}  & \checkmark & -   & 76.8 &  65.6 \\
        SAM2~\cite{ravi2024sam2}  & \checkmark & \checkmark   & 77.5 &  63.0 \\
        E-RADIO~\cite{ranzinger2024amradio}  & \checkmark & \checkmark & 78.9 &  68.2 \\
        DINOv2~\cite{dinov2_2023}              & \checkmark & -  & 79.1 &  68.0  \\ 
        \rowcolor{gray!30}
        DINOv2~\cite{dinov2_2023}~+~ViT-Adapter~\cite{chen2022vitadapter} & \checkmark & \checkmark  & \textbf{80.3} &  \textbf{68.6} \\ 
    \bottomrule
    \end{tabular}
\end{adjustbox}
\caption{\label{tab:vfmpretraining} \textbf{Ablation of different single-scale and multi-scale VFMs.} DINOv2 paired to a ViT-Adapter provides optimal generalization and up to +1.4 mIoU better than the best multi-scale VFM E-RADIO. Using the All-synth to All-real benchmark with ViT-B/14 VFM-UDA++ model size for all metrics. }
\vspace{-5pt} 
\end{table}

\PAR{Ablation of multi-scale decoder:}
For multi-scale decoder configurations, we evaluate \textit{BasicPyramid}, DAFormer~\cite{hoyer_daformer_2022}, HRDA~\cite{hoyer_hrda_2022}, and Mask2Former~\cite{cheng2021mask2former} in conjunction with the ViT-Adapter. As shown in ~\cref{tab:multiscale}, \textit{BasicPyramid} achieves the highest mIoU performance with the most favorable inference time, showing that a more complicated decoder is unnecessary for UDA.  Surprisingly, more complex decoders such as Mask2Former, which incorporate transformer-based mask prediction and multi-stage refinement, did not yield better results. While HRDA (also used by the recent MIC~\cite{mic_2023}) and Mask2Former reach similar mIoU scores (80.3\% and 80.0\%, respectively), they do so at a significant computational cost, requiring 5.09$\times$ and 3.96$\times$ more inference time, respectively.

These findings suggest that \textit{BasicPyramid} strikes the best balance between performance and efficiency when paired with the ViT-Adapter. It delivers strong segmentation accuracy with minimal computational overhead compared to other decoder heads, making it the preferred architecture choice for VFM-UDA++.

\begin{table}[tb]
\centering 
\begin{adjustbox}{width=\linewidth}
\begin{tabular}{cccccccc}
\toprule
\thead{\textbf{Encoder} \\ \textbf{modification}} & \textbf{Decoder} & \thead{\textbf{Multi-scale} \\ \textbf{encoder}} & \thead{\textbf{Multi-scale} \\ \textbf{decoder}} & $\#$\textbf{Param} & \thead{\textbf{Inference} \\ \textbf{time}} & \textbf{mIoU CS} \\
\midrule
-- & VFM-UDA~\cite{englert2024exploring} & -- &  -- & 88M & 1.00 $\times$ & 79.1  \\ %
ViT-Adapter~\cite{chen2022vitadapter} & VFM-UDA~\cite{englert2024exploring} &  $\checkmark$ &  -- & 102M & 1.81 $\times$ & 79.9 \\ %
2 Extra ViT Blk. & VFM-UDA~\cite{englert2024exploring} & -- &  -- & 102M & 1.05 $\times$ & 78.8 \\  %
\midrule 
\rowcolor{gray!30}
ViT-Adapter~\cite{chen2022vitadapter} & BasicPyramid &  $\checkmark$ &  $\checkmark$ & 104M & 1.80 $\times$ & \textbf{80.3} \\ %
ViT-Adapter~\cite{chen2022vitadapter} & DAFormer~\cite{hoyer_daformer_2022} &  $\checkmark$ &  $\checkmark$ & 104M & 2.14 $\times$ & 79.7 \\ %
ViT-Adapter~\cite{chen2022vitadapter} & HRDA~\cite{hoyer_hrda_2022} &  $\checkmark$ &  $\checkmark$ & 109M & 5.09 $\times$ &  80.3 \\ %
ViT-Adapter~\cite{chen2022vitadapter} & Mask2Former~\cite{cheng2021mask2former} &  $\checkmark$ &  $\checkmark$ & 120M &  3.96 $\times$  & 80.0 \\ %
\bottomrule
\end{tabular} 
\end{adjustbox}
\caption{\textbf{Ablation of multi-scale encoder and decoder architectures with the \textit{All-synth $\rightarrow$ All-real} benchmark.} 
The results show the optimal mIoU vs. Inference time of using ViT-Adapter with our custom \textit{BasicPyramid} decoder.
The inference time is measured on an Nvidia H100 GPU with 16-bit mixed precision for ViT-B/14 model size.}
\label{tab:multiscale}

\vspace{-15pt} 
\end{table}

\subsection{Effect of Feature Distance Loss}
\label{sec:ablation_train_strat} Shown in ~\cref{tab:fdistanceloss}, we evaluated the effect of added feature distance (FD) losses on top of the VFM-UDA training strategies, comparing DaFormer’s feature distance loss to our modern VFM counterpart, which is not used before in the context of UDA. The VFM feature distance loss benefit is more pronounced in small-scale data settings where it adds +1.0 mIoU over VFM-UDA. This shows that a modern version of feature distance loss can help UDA performance when using ViT based VFMs. In contrast, the DAFormer~\cite{hoyer_daformer_2022} feature distance loss used in SotA non-VFM methods decreases performance with -4.1 mIoU.

\begin{table}[tb]
\centering 
\begin{adjustbox}{width=1.0\linewidth}
\begin{tabular}{cccccl}
\toprule 
\textbf{Source} & \textbf{Target} & \textbf{Baseline} & \thead{\textbf{w/ DaFormer's} \\ \textbf{FD loss}~\cite{hoyer_daformer_2022}} &  \thead{\textbf{w/  VFM}  \\ \textbf{FD loss}} & \\
\midrule
GTA5 & CS & 78.1 & 74.0~\reddown~\textcolor{BrickRed}{-4.1} & \textbf{79.1}~\greenup~\textcolor{applegreen}{+1.0} & mIoU CS\\
All-synth & All-real &  80.3 & 79.9~\reddown~\textcolor{BrickRed}{-0.4} & \textbf{80.6}~\greenup~\textcolor{applegreen}{+0.3} & mIoU CS\\
\bottomrule
\end{tabular}
\end{adjustbox}
\caption{\textbf{Relevance of feature distances losses in VFM-UDA++} We show that the VFM feature distance loss yields the best performance, while DaFormer's feature distance loss decreases it slightly. Using the All-in benchmark with ViT-B/14 VFM-UDA++ architecture for all metrics.} 
\label{tab:fdistanceloss}
\vspace{-15pt} 
\end{table}

\begin{figure*}
    \centering
    \includegraphics[width=0.98\linewidth]{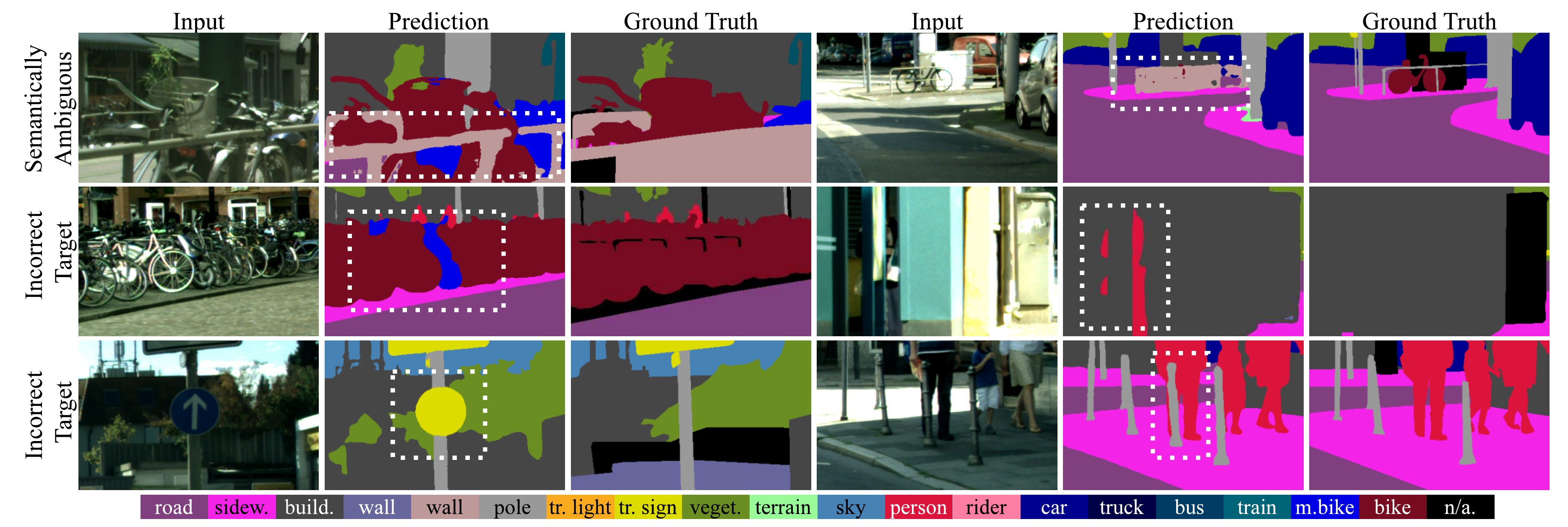}
    \caption{\label{fig:qualitative} \textbf{Predictions by VFM-UDA++ can surpass ground truth accuracy}. Best viewed in high resolution. We observe that due to semantic ambiguity and manual labeling errors, the predictions of VFM-UDA++ can be more accurate than the ground truth. \textbf{Top left}: The model correctly predicts individual bars of the fence, while these are not labeled in the ground truth. \textbf{Top right}: The model predicts the entire fence, while now the individual bars are labeled. \textbf{Middle left}: The model correctly predicts a motorbike that is labeled as a bike. \textbf{Middle right}: The model correctly predicts a human that is not labeled. \textbf{Bottom left}: The model correctly predicts a traffic sign that is not labeled. \textbf{Bottom right}: The model correctly predicts a pole that is not labeled. }
 \vspace{-15pt} 
\end{figure*}

\subsection{Effect of Large-Scale Datasets} %
\label{diversedatacombinations}

\PAR{Data Scalability of UDA:} In \cref{tab:scalingdata}, we report results on the \textit{All-synth} $\rightarrow$ \textit{All-real} benchmark and analyze how generalization scales with data. VFM-UDA and VFM-UDA++ both show consistent improvements as more diverse synthetic and unlabeled real data is introduced. In contrast, the non-VFM UDA method MIC degrades in performance when trained with more data, highlighting a key difference in scalability between VFM- and non-VFM-based methods.

For VFM-UDA++, scaling from standard synthetic datasets to the full \textit{All-synth}, and from Cityscapes-only to the \textit{All-real} target set, yields a +2.4 MIOU gain, reaching \textbf{82.2 mIoU}. When compared to the fully-supervised oracle (85.0 mIoU), this leaves a gap of only 2.8 mIoU, demonstrating that VFM-pretrained models are well-suited to benefit from larger-scale UDA setups.

\PAR{Influence of Dataset Composition:} While combining large-scale synthetic and real target data proves effective overall, the individual contributions of each component are not immediately obvious. Understanding this composition is crucial for applying UDA effectively in practice. In \cref{tab:dataablation}, we begin by isolating and evaluating the contribution of individual synthetic datasets to the source-only generalization performance of VFM-based UDA.

Individually, the synthetic datasets are relatively weak: Synthia achieves only 46.4 mIoU, Synscapes 59.7, and GTA5 59.2. These results highlight that no single dataset is sufficient to close the domain gap. Interestingly, when the synthetic datasets are combined, source-only generalization, without any target images, already reaches 72.2 mIoU. This is  higher than some prior UDA methods reported in \cref{tab:gta2cs}. This result suggests that strong performance can be obtained solely through better architectures, VFM-based pretraining, and a richer source domain alone, even without target-domain adaptation.

Naturally, performance improves further with UDA when target data is introduced. The bottom half of \cref{tab:dataablation} evaluates various combinations of source and target data. A key observation emerges when we look at in-target generalization to Cityscapes (CS). The highest performance (79.3 mIoU) is achieved with \textit{All-synth} $\rightarrow$ CS. Adding more target variety, as in \textit{All-synth} $\rightarrow$ \textit{All-real}, reduces this to 73.4 mIoU. Even smaller configurations like GTA5 $\rightarrow$ CS (77.1 mIoU) outperform GTA5 $\rightarrow$ \textit{All-real} (73.4 mIoU). This trend is consistent with the trend observed for MIC in \cref{tab:scalingdata}, where broadening the target domain hurts the performance of a specific, narrow target domain.

However, this effect can be partially mitigated. When we combine diverse synthetic sources (\textit{All-synth}), the performance drop from using broader target data is smaller (78.6 mIoU vs. 73.4 mIoU). Moreover, while broader target data may hurt in-target generalization to Cityscapes, it consistently improves performance on out-of-target evaluations like WildDash2. This reinforces the idea that broader real target data aids generalization across diverse domains.

From this, we can derive several important recommendations. Scaling labeled source data is a relatively safe approach to improving the generalization of UDA. When scaling the target data, one needs to be more careful, as adding too much variety can also hurt performance to a specific narrow target. It is highly advisable to scale source data when scaling target data, as it can compensate for the negative effect of too varied target data and help to generalize.

\begin{table}[tb] 
\centering 
\begin{adjustbox}{width=\linewidth}
\begin{tabular}{lccccc} 
\toprule
\textbf{Method}  & \textbf{$\#$Parameters} & \thead{\textbf{GTA5 $\rightarrow$ Cityscapes} \\ mIoU CS} & \thead{\textbf{All-synth $\rightarrow$ All-real} \\ mIoU CS} \\
\midrule
MIC B~\cite{mic_2023}              &  86M  & 75.9 & 73.7~\reddown~\textcolor{BrickRed}{-2.2} \\
\midrule
VFM-UDA S~\cite{englert2024exploring}          &  27M  & 71.3 & 74.8~\greenup~\textcolor{applegreen}{+3.5} \\
VFM-UDA B~\cite{englert2024exploring}          &  88M  & 77.1 & 79.1~\greenup~\textcolor{applegreen}{+2.0} \\
VFM-UDA L~\cite{englert2024exploring}          &  334M & 78.4 & 80.8~\greenup~\textcolor{applegreen}{+2.4} \\
\midrule
\textcolor{gray}{Source-only} &  \textcolor{gray}{334M}  & \textcolor{gray}{71.8} & \textcolor{gray}{75.9} \\
VFM-UDA++ S &  27M  & 76.6 & \textbf{77.9}~\greenup~\textcolor{applegreen}{+1.3} \\
VFM-UDA++ B &  88M  & 79.1 & \textbf{80.6}~\greenup~\textcolor{applegreen}{+1.5} \\
VFM-UDA++ L &  334M & 79.8 & \textbf{82.2}~\greenup~\textcolor{applegreen}{+2.4} \\
\textcolor{gray}{Oracle} &  \textcolor{gray}{334M}  &  \textcolor{gray}{85.0}    & \textcolor{gray}{85.0} \\
\bottomrule
\end{tabular}
\end{adjustbox} 
\caption{\label{tab:scalingdata} \textbf{All-synth $\rightarrow$ All-real benchmark.}VFM-UDA and VFM-UDA++ are able to scale with data obtaining improved generalization while MIC is not. The provided deltas are to the standard GTA5 $\rightarrow$ Cityscapes results. } 

\vspace{-15pt} 
\end{table}

\begin{table}[tb]
\centering
\begin{adjustbox}{width=0.70\linewidth}
\begin{tabular}{cccc}
\toprule
\textbf{Source} & \textbf{Target} & \thead{\textbf{In-target} \\  \textbf{mIoU CS}}  & \thead{\textbf{Out-of-target} \\  \textbf{mIoU WD2}} \\
\midrule
 Synthia   & - & 46.4 & 33.1 \\
 Synscapes & - & 59.7 & 45.5 \\
 GTA5      & - & 59.2 & 53.7 \\
 All-synth & - & \textbf{72.2} & \textbf{58.8} \\
 \midrule
 GTA5      & Cityscapes & 77.1 & 61.3 \\ 
 All-synth & Cityscapes & \textbf{79.3} & 61.8 \\
 GTA5      & All-real   & 73.4 & 63.0 \\
 All-synth & All-real   & 78.6 & \textbf{67.1} \\
\bottomrule
\end{tabular}
\end{adjustbox}
\caption{\label{tab:dataablation} \textbf{Data combination ablations}. Showing that diversity in source data is beneficial, while diversity for real target data can harm in-target, but improve out-of-target performance.}

\vspace{-15pt} 
\end{table}

\subsection{Out-of-Target Generalization}
In this experiment, we focus on the setting where the exact distribution of the target domain is unknown beforehand This is relevant to many real-world UDA applications where a computer vision model can be deployed in an environment that cannot be fully anticipated beforehand. The results in \cref{tab:outoftarget} show the out-of-target mIoU of UDA methods on WildDash2 (WD2) when increasing data from GTA5, to GTA5 and Cityscapes, and finally \textit{All-synth} and \textit{All-real}. For the out-of-target generalization we see a positive trend for all non-VFM and VFM UDA methods, in contrast to the in-target results in \cref{tab:scalingdata}. The results in \cref{tab:scalingdata} and \cref{tab:outoftarget} show that the more diverse source and target data of \textit{All-synth $\rightarrow$ All-real} help UDA to generalize better to the broader out-of-target distribution of WildDash2, but not necessarily to the narrower in-target of Cityscapes.

\begin{table}[tb]
\centering
\begin{adjustbox}{width=0.7\linewidth}
\begin{tabular}{lccc}
\toprule

 \textbf{Method} & \textbf{Source} & \textbf{Target}  & \thead{\textbf{Out-of-target} \\  \textbf{mIoU WD2}} \\
\midrule
\textcolor{gray}{Source-only} & \textcolor{gray}{GTA5} & \textcolor{gray}{-} & \textcolor{gray}{40.6} \\
MIC~\cite{mic_2023} & GTA5      & CS       & 55.2~\greenup~\textcolor{applegreen}{+4.6} \\
MIC~\cite{mic_2023} & All-synth & All-real & 60.1~\greenup~\textcolor{applegreen}{+4.9} \\
\midrule
\textcolor{gray}{Source-only} & \textcolor{gray}{GTA5} & \textcolor{gray}{-} & \textcolor{gray}{64.8} \\
VFM-UDA~\cite{englert2024exploring} & GTA5      & CS       & 65.5~\greenup~\textcolor{applegreen}{+0.7} \\
VFM-UDA~\cite{englert2024exploring} & All-synth & All-real & 70.4~\greenup~\textcolor{applegreen}{+4.9} \\
\midrule
\textcolor{gray}{Source-only} & \textcolor{gray}{GTA5} & \textcolor{gray}{-} & \textcolor{gray}{65.5} \\
VFM-UDA++   & GTA5      & CS       & 69.0~\greenup~\textcolor{applegreen}{+3.5} \\
VFM-UDA++   & All-synth & All-real & \textbf{71.3}~\greenup~\textcolor{applegreen}{+2.3} \\
\bottomrule
\end{tabular}
\end{adjustbox}
\caption{\label{tab:outoftarget} \textbf{Out-of-target performance when scaling data}. All UDA methods scale with data for out-of-target generalization. The provided deltas are to the preceding row in the table.}
\vspace{-15pt}
\end{table}

\subsection{Comparing VFM-UDA++ to Oracle Models}
In this experiment, we compare the generalization performance of VFM-UDA++ to several supervised oracle models to contextualize its performance clearly. First, we compare VFM-UDA++ to the current state-of-the-art on Cityscapes, to examine how competitive the architecture itself in a fully-supervised setting. Second, we evaluate synthetic source-only supervised learning approaches, such as Rein~\cite{wei2024stronger}. Finally, we introduce a \textit{cross-domain oracle}, trained on the \textit{All-real} dataset excluding Cityscapes, providing a meaningful reference point for generalization limits without direct target-domain annotations. The rationale and definition of the cross-domain oracle are described in detail in \cref{sec:exp_setup}, and further illustrated qualitatively in \cref{fig:qualitative}, which highlights annotation inconsistencies and labeling errors in Cityscapes.

The standard oracle trained on Cityscapes using the VFM-UDA++ architecture reaches 85.0 mIoU. In comparison, larger models such as ViT-Adapt~\cite{chen2022vitadapter} + M2F~\cite{cheng2021mask2former} (571M) and MetaPrompt-SD~\cite{wan2023harnessing} (912M) only slightly improve performance, reaching 85.8 and 86.0 mIoU respectively. This suggests that further scaling of model size yields diminishing returns in this benchmark, and shows that VFM-UDA++ has a competitive architecture for semantic segmentation.

Source-only supervised models trained on synthetic datasets lag significantly behind. For instance, Rein~\cite{wei2024stronger} achieves 66.4 mIoU when trained on GTA5 alone, and improves to 78.4 mIoU when including additional synthetic datasets (GTA, Synthia, and USeg). Despite these improvements, the gap to UDA methods, which can leverage unlabeled real data, remains substantial.

The cross-domain oracle, introduced in \cref{sec:exp_setup}, is trained on all labeled real-world datasets except Cityscapes and evaluated on the Cityscapes validation set. It achieves 83.0 mIoU, 2.0 mIoU points below the standard oracle trained directly on Cityscapes. As discussed earlier, this large gap may reflect not only domain shift but also label inconsistencies between the datasets. As such, the cross-domain oracle provides a complementary reference point that better accounts for dataset mismatch and labeling quality in UDA evaluation.

To contextualize this, VFM-UDA++ achieves 82.2 mIoU when trained using the full All-synth  All-real UDA setup, closing the gap to the cross-domain oracle to just 0.8 mIoU. This demonstrates that UDA performance can come close to fully-supervised learning under certain conditions, even without using target-domain labels.

\begin{table}[tb]
\centering
\begin{adjustbox}{width=1\linewidth}
\begin{tabular}{lccccccc}
\toprule
\textbf{Architecture} & \textbf{Size} & \textbf{Setting} & \textbf{Source} & \textbf{Target} & \textbf{mIoU CS} \\
\midrule
VFM-UDA++           &  334M  & UDA      & GTA5        & CS       & 79.8   \\
VFM-UDA++           &  334M  & UDA      & GTA5,S,US   & CS       & 81.5   \\
\rowcolor{gray!30}
VFM-UDA++           &  334M  & UDA      & All-synth   & All-real & \textbf{82.2}   \\
\midrule
VLTSeg~\cite{Hummer2024}    & 304M  & Source-only      & GTA5            & - & 65.3  \\
Rein~\cite{wei2024stronger} & 350M  & Source-only      & GTA5            & - & 66.4  \\ 
Rein~\cite{wei2024stronger} & 350M  & Source-only      & GTA,S,US        & - & 78.4  \\
\rowcolor{gray!30}
VFM-UDA++ (oracle)  & 334M  &  Source-only       & All-real w/o CS & - & \textbf{83.0} \\
\midrule
\rowcolor{gray!30}
VFM-UDA++ (oracle)  & 334M  & fully-sup.       & CS              & - & 85.0 \\
ViT-Adapt~\cite{chen2022vitadapter}~+~M2F~\cite{cheng2021mask2former}     & 571M  & fully-sup.       & CS              & - & 85.8 \\
MetaPrompt-SD~\cite{wan2023harnessing}       & 912M  & fully-sup.       & CS              & - & \textbf{86.0} \\
\bottomrule
\end{tabular}
\end{adjustbox}
\caption{\label{tab:supervised}\textbf{Oracle performance comparisons.} The table compares VFM-UDA++ to fully-supervised oracles and source-only models, showing how different supervision levels and data combinations affect performance on Cityscapes.}
\vspace{-15pt} 
\end{table}

\section{Conclusions}
\label{sec:conclusions}
This work investigated how Unsupervised Domain Adaptation (UDA) can benefit from Vision Foundation Models (VFMs). We proposed VFM-UDA++, which extends prior VFM-based UDA methods by reintroducing multi-scale inductive biases, adapting feature distance losses for ViT-based encoders, and evaluating the effects of scaling synthetic source and unlabeled target data.
Our experiments show that multi-scale features can improve performance, but not all multi-scale encoder configurations were beneficial. In fact, the best results were achieved using a single-scale ViT backbone pretrained with DINOv2, with further gains from augmenting it with a ViT-Adapter. Similarly, decoder design proved important, many off-the-shelf decoders were either too slow or underperformed; a lightweight, well-matched decoder was most effective. Feature distance losses, which were previously ineffective in VFM-based UDA settings, become effective in VFM-UDA++ by removing class-based masking and enabling label-free application on both source and target domains. Finally, we examined the effect of data scale, showing that VFM-UDA++ benefits from both larger synthetic sources and more diverse unlabeled target data. However, this effect is not uniform: while broader target data improves out-of-domain generalization, it can hurt performance on narrow in-target distributions like Cityscapes unless compensated by scaling the source domain as well. 
Taken together, our results show that meaningful gains in UDA can be achieved through careful adaptation of architectures, loss functions, and data composition, rather than simply scaling models or relying on generic components. These insights are especially relevant for applications like autonomous driving, where models must generalize reliably across varied, unlabeled real-world conditions.

\section*{Acknowledgements}
This work was funded by the Horizon Europe programme of the European Union, under ChipsJU grant agreement No.101097300 (project EdgeAI).

The project Ecomobility has received funding within the Chips Joint Undertaking (Chips JU), a public-private partnership in collaboration with the Horizon Europe (HORIZON) Framework Programme and National Authorities under grant agreement number 101112306.

This work made use of the Dutch national e-infrastructure with the support of the SURF Cooperative, using grant no. EINF-10932,  EINF-11473 and EINF- 7181, which is financed by the Dutch Research Council (NWO).

\section*{Disclaimer}
Funded by the European Union. Views and opinions expressed are however those of the author(s) only and do not necessarily reflect those of the European Union or the Chips Joint Undertaking. Neither the European Union nor the granting authority can be held responsible for them.

{
    \small
    \bibliographystyle{ieeenat_fullname}
    \bibliography{main}
}

\end{document}